# Evolution of Biped Walking Using Neural Oscillators Controller and Harmony Search Algorithm Optimizer

Ebrahim Yazdi, Abolfazl Toroghi Haghighat

**Abstract**— In this paper, a simple Neural controller has been used to achieve stable walking in a NAO biped robot, with 22 degrees of freedom that implemented in a virtual physics-based simulation environment of Robocup soccer simulation environment. The algorithm uses a Matsuoka base neural oscillator to generate control signal for the biped robot. To find the best angular trajectory and optimize network parameters, a new population-based search algorithm, called the Harmony Search (HS) algorithm, has been used. The algorithm conceptualized a group of musicians together trying to search for better state of harmony. Simulation results demonstrate that the modification of the step period and the walking motion due to the sensory feedback signals improves the stability of the walking motion.

——————————— ◆ ———————————

## 1 INTRODUCTION

The infrastructure of our society is designed for humans. For example, the sizes of doors and the heights of steps on stairs are determined by considering the heights of people and the lengths of their legs. Therefore, we can apply robots for our society without extra investment in the infrastructure if the robots have the human shape [1]. In researches about bipedal robots, walking is one of the main challenges. There are two major approaches, model-based and model-free, in bipedal walking researches. In model-based approaches, controller of robot is dependent on model of robot and from one robot to another every thing in controller should be changed. Two well known methods in this approach are "Zero Moment Point"[2, 3] (ZMP) and "Inverted Pendulum"[4]. The animated end results of these efforts closely resemble natural motion patterns, but may nevertheless fail to convince the human eye in specifically designed "motion Turing tests" [5]. This lack of realism is a direct consequence of the controller architecture employed; a state machine does not readily produce the fluctuations typical of real locomotion. More significantly, it cannot easily be extended to integrate sensory input. Furthermore, the creation of state machines can be a cumbersome process, as states have to be identified, implemented and fine-tuned by hand for each type of gait to be modeled [5]. In model-free approach, controller of robot is independent of its model. Model-free approach has two portions. A portion for control of the robot and a portion for find the best values for variables of controller. Genetic Algorithm Optimized Fourier Series Formulation (GAOFSF) [6] as a model free approach is imitated of Human's walk.

GAOFSF was used in 2006 for the first time for gait generation in bipedal locomotion [7]. GAOFSF uses TFS (Truncated Fourier Series; =modified definite Fourier serie equation) to control the robot and uses from genetic algorithms for optimizing the parameters of TFS. Central Pattern Generator (CPG) as a model-Free approach uses from a set of neural oscillators for controlling and uses genetic algorithm as a weight optimizer. Many studies have been carried out in order to elucidate the role for CPG in locomotion using quadruped robots [8], [9], biped robots [10], [11], [12], and simulated salamander [13].

In this paper, we investigate experimentally sensory feedback neural network controller for biped locomotion based on Matsuoka [14] formulate. A controller composed of nonlinear oscillators receives feedback signals from touch sensors. Instantly, feedback sensors send signals to this network, its neural oscillators generate output signals for applying to joint angle trajectories of robot.

In similar works[15,16] GA used as optimizer for finding best weights for neural networks, but in this approach, the Harmony Search (HS) algorithm [17] technique with constraint handling on angles and time is used to find optimum weights of network and train the robot to achieve fast bipedal walking, for the first time. HS is a meta-heuristic algorithm, mimicking the improvisation process of music players [18]. During last several years, the HS algorithm has been vigorously applied to various optimization problems [18, 19], presenting several advantages with respect to traditional optimization techniques such as the following [18]: (a) HS algorithm is simple in concept, few in parameters, easy in implementation, imposes fewer mathematical requirements, and does not require initial value settings of the decision variables. (b)

————————————————

- *E.Yazdi is with the Department of Computer Engineering & IT, Qazvin Azad University, Qazvin, Iran*
- *A.Toroghi Haghighat. is with the Department of Computer Engineering & IT, Qazvin Azad University, Qazvin, Iran*

As the HS algorithm uses stochastic random searches, derivative information is also unnecessary. (c) The HS algorithm generates a new vector, after considering all of the existing vectors, whereas the GA only considers the two parent vectors. These features increase the flexibility of the HS algorithm and produce better solutions.

The paper is organized as follows: Section 2 provides a definition about the simulator, Nao robot and explanation about important joints for walking. In section 3, we provide a formal definition of the Matsuoka formulate and using this to our suggested Neural Network. Section 3, begins with explanations about HS and continues with applying HS as an optimizer to our algorithm. Section 4 gives the performance evaluation of the proposed algorithm compared to GA-Based algorithm. Section 5 concludes the paper.

## 2 SIMULATOR AND ROBOT MODEL

The target Robot of our study is a 22-DOF (degrees of freedom) NAO robot (Fig.1) with 4.5Kg weight and 57Cm stand height. The robot has two arms with four DOF for each arm, two legs with six DOF for each leg and a head with 2 degrees of freedom.

The simulation performed by Rcssserver3d simulator which is a generic three dimensional simulator based on Spark and Open Dynamics Engine (ODE [20]). Spark is capable of carrying out scientific distributed multi agent calculations as well as various physical simulations ranging from articulated bodies to complex robot environments [21]. The time-integrated simulation is processed with a resolution of 50 simulation steps per second. Rcssserver3d simulator is a noisy simulator and to overcome inherent noise of the simulator, Resampling algorithm is implied which could lead to robustness in nondeterministic environments.

In this approach, the body trunk of the robot is not actuated. Our experiences show, the leg joints are more effective joints for leg motions, that the joints of hip, knee and ankle which move on the same plane of forward-backward are the major ones (Table 1 shows description of them). Although, other joints are also effective, but in fact, their role smoothes the Robot walking motion. Because of this we prefer to ignore them to decrease learning search space.

## 3 NEURAL SYSTEM MODEL

### 3.1 Matsuoka Neural oscillator

Matsuoka oscillator is a kind of relaxation oscillator and is basically composed of a pair of mutually inhibiting neural elements with adaptation effects (i.e. exhaustion after excitation) [14].

The original Matsuoka neural oscillator formulates can be written as follow:

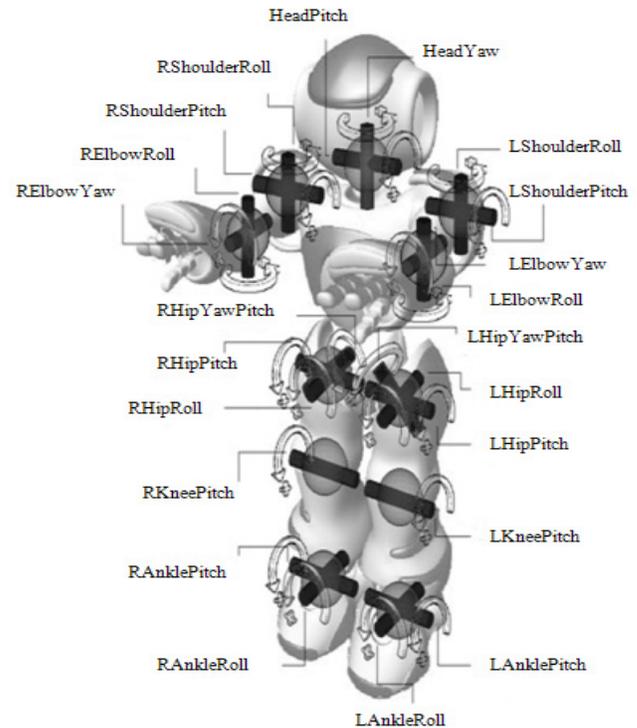

Fig. 1. The biped physical structure and its twenty two degrees of freedom. This figure indicates the joint angle and shows the robot coordinate system..

TABLE 1
MORPHOLOGICAL MAJOR PARAMETERS OF THE NAO ROBOT

| Joint Name | Motion | Range (degree) |
|---|---|---|
| *Right Leg* | | |
| RHipPitch | Right hip joint front & back (Y) | -100 to 25 |
| RKneePitch | Right knee joint (Y) | -130 to 0 |
| RAnklePitch | Right ankle joint front & back (Y) | -75 to 55 |
| *Left Leg* | | |
| LHipPitch | Left hip joint front & back (Y) | -100 to 25 |
| LKneePitch | Left knee joint (Y) | -130 to 0 |
| LAnklePitch | Left ankle joint front & back (Y) | -75 to 55 |

$$\tau_1 \dot{x}_1 = c - x_1 - \beta v_1 - \gamma [x_2]^+ - \alpha u_{f1} \quad (1)$$

$$\tau_2 \dot{v}_1 = [x_1]^+ - v_1 \quad (2)$$

$$\tau_1 \dot{x}_2 = c - x_2 - \beta v_2 - \gamma [x_1]^+ - \alpha u_{f2} \quad (3)$$

$$\tau_2 \dot{v}_2 = [x_2]^+ - v_2 \quad (4)$$

Where $x_1$, $x_2$, $v_1$, $v_2$ are internal variables; $\tau_1$, $\tau_2$ are time constants; $\beta$ represents the intensity (suppression) of adaptation, $c$ means the constant input; $u_{f1}$, $u_{f2}$ are the feedback inputs mainly from sensors, the inputs scaled by gain $\alpha$; $\gamma$ is the connection coefficient of two exhaustive elements. Given a certain constant input ($c$) to both of the two elements, either of the elements with higher state values initially gets exited more than the other and the more excited one inhibits the other. As a result, the excited element sends the positive feedback to the other element and because of receiving inhabitation from the other element becomes weaker. Then after the excitation of one element reaches the peak, it gradually relaxes due to the adaptivity. This weakens the suppression to the other element. As a result the excitation of the other element grows stronger. So, we have a symmetrical opposite signals to the initial state and this process continues like seasaw game. If $uf1, uf2$ are large enough, and close to the oscillator's natural frequency, the phase difference between input and output is tightly locked due to entrainment. Figure 2 shows this neural oscillator architecture.

### 3.2 THE NEURAL ARCHITECTURE

Human motions are recognized flexible and periodic but more challenging for motion stability issue. Therefore, human-like motion patterns are included into the re- search objectives. The walking trajectory is divided into several types. Positional trajectory and angular trajectory are two of them. The main idea for the walking posture control used here is based on the reaction angular trajectory. Similar to [22], Foot was kept parallel to the ground by using ankle joint in order to avoid collision. Therefore, ankle trajectory can be calculated by hip and knee trajectories and ankle DOF parameters are eliminated.

According to this fact that human walking is a periodic motion and attention to this point that Matsuoka oscillator can generate periodic signals, our neural oscillator approach generates a core oscillation with the use of the Matsuoka oscillator. One of the elements in the Matsuoka oscillator corresponds to left leg and the other to right leg. The out put of the Matsuoka oscillator given to same leg neurons. Besides, We have taken a neuron for each joint angles of left hip, right hip, left knee, right knee (the activity function of these neurons is PureLin) and the output of each Matsuoka elements, By multiplying by a separate weight would be exerted to the neurons in charge. Multiplying the weight in matsuoka`s signals causes to change the output wave length to each neuron and turn to the quantity required for each joint. Eventually, A bios has been assigned to every Purelin`s neurons. This bios can shift the location of the generated signals in the degree axis of each joint to the negative or positive direction (on the other hand, to add a permanent offset as the same quantity of the input signal of every joint). Attention to this point that researchers are always looking for small the search space, bios and weight in each one of left hip-right hip and left knee-right knee neurons can be chosen the same separately (because the movement of each of these joints is isochronous and they have just half of the period difference at the time of occurrence) that this can cause the search space to be smaller. after exerting achieved results from neural oscillators on the robot joints, left hip and right hip joint angles would be exerted to Matsuoka`s neurons that is responsible for generating their signals respectively.

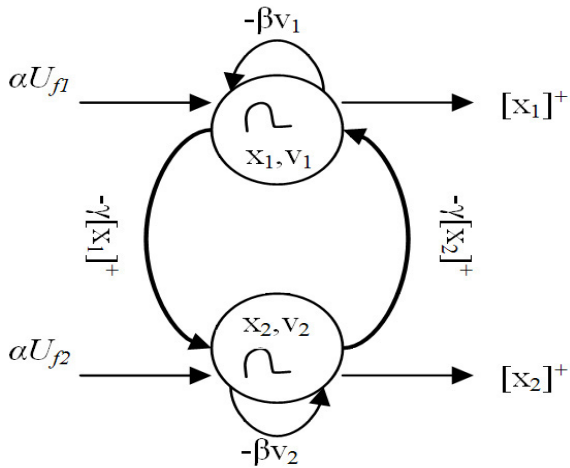

Fig. 2. Matsuoka Coupled Neural Oscillator. When a constant input is fed to an adaptive neural element, its responsive output first approaches the constant input, but then after some point the output declines because of the adaptivity of the element.

## 4 HARMONY SEARCH ALGORITHM

### 4.1 Harmony Search Algorithm Definition

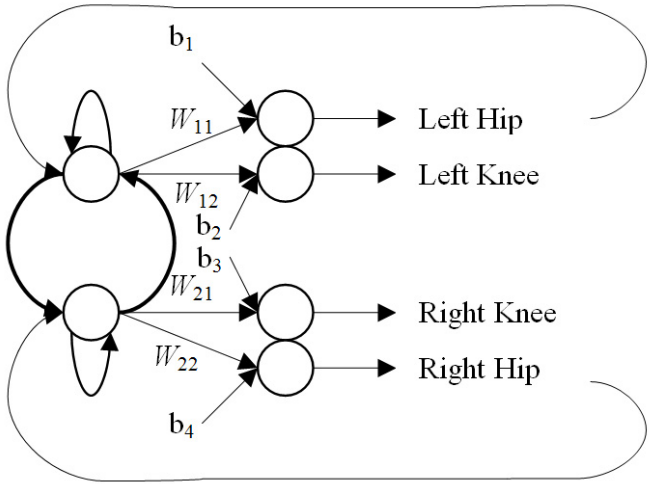

Fig. 3. Neural architecture: Input layer is a Matsuoka coupled neural oscillator. Neurons of layer one are PureLin. In this network $w_{11}=w_{22}$, $w_{12}=w_{21}$, $b_1=b_4$ and $b_2=b_3$. Results of Left Hip and Right Hip back to input layer.

Harmony Search (HS) algorithm which is a nature-inspired algorithm, mimicking the improvisation process of music players has been recently developed by Geem et al. [18]. The HS algorithm is simple in concept, few in parameters, and easy in implementation, it has been successfully applied to various benchmarking and real-world problems including travelling salesman problem [19]. The main steps of HS are as follows [19]:

Step 1: Initialize the problem and algorithm parameters.

Step 2: Initialize the harmony memory.

Step 3: Improvise a new harmony.

Step 4: Update the harmony memory.

Step 5: Check the stopping criterion.

These steps are described in the next five subsections.

### 4.1.1 Initialize the Problem and Algorithm Parameters

As demonstrated in this document, the numbering for sections upper case Arabic numerals, then upper case Arabic numerals, separated by periods. Initial paragraphs after the section title are not indented. Only the initial, introductory paragraph has a drop cap.

In the first step, the optimization problem is specified as follows:

Minimize (or Maximize) $f(x)$ Subject to,
$$x_i \in X_i, i = 1,2,...,N \qquad (5)$$
Where $f(x)$ is an objective function; $x$ is a solution vector composed of each decision variable $x_i$. $X_i$ is the set of possible range of values for each decision variable, that is, $x_i=\{x_{i1},x_{i2},...,x_{ik}\}$ for discrete decision variables (Generally, candidate values are sorted like $x_{i1}<x_{i2}<...<x_{ik}$); $N$ is the number of decision variables and $K$ is the number of candidate values for the discrete decision variables. The HS algorithm parameters are also specified in this step. These are harmony memory size (HMS) or number of simultaneous solution vectors in harmony memory, harmony memory considering rate (HMCR), pitch adjusting rate (PAR) and number of improvisations (NI). The harmony memory (HM) is a memory location where all the solution vectors (sets of decision variables) are stored. This HM is similar to the genetic pool in the GA.

### 4.1.2 Initialize the Harmony Memory

In Step 2, the Harmony Memory (HM) as shown in Equation 7, is crammed with as many randomly generated solution vectors as the size of the HM (i.e., HMS).

$$\mathbf{HM} = \begin{bmatrix} x_1^1 & x_2^1 & \cdots & x_N^1 \\ x_1^2 & x_2^2 & \cdots & x_N^2 \\ \vdots & \vdots & \vdots & \vdots \\ x_1^{HCM} & x_2^{HCM} & \cdots & x_N^{HCM} \end{bmatrix} \qquad (6)$$

### 4.1.3 Improvise a New Harmony

A New Harmony vector $x' = (x'_1, x'_2, \cdots, x'_N)$ is improvised by following three rules: (1) random selection, (2) HM consideration, and (3) pitch adjustment. Generating a new harmony is called 'improvisation'.

In the memory consideration, the value of the first decision variable ($x'_1$) for the new vector is chosen from any of the values in the specified HM range ($x'_1{}^1 - x'_1{}^{HMS}$). Values of the other decision variables ($x'_2, x'_3, ..., x'_N$) are chosen in the same manner. The HMCR, which varies between 0 and 1, is the rate of choosing one value from the historical values stored in the HM, while (1 _ HMCR) is the rate of randomly selecting one value from the possible range of values.

$$x'_i \leftarrow \begin{cases} x'_i \in \{x_i^1, x_i^2, \cdots, x_i^{HMS}\} & \text{with probability } HMCR \\ x'_i \in X_i & \text{with probability } (1-HMCR) \end{cases} \qquad (7)$$

Every component obtained by the memory consideration is examined to determine whether it should be pitch-adjusted. This operation uses the PAR parameter, which is the rate of pitch adjustment as follows:

$$\text{pith Adjusting Decision for } x'_i \leftarrow \begin{cases} \text{Yes with probability } par \\ \text{No with probability } (1-par) \end{cases} \qquad (8)$$

The value of (1 - PAR) sets the rate of doing nothing. If the pitch adjustment decision for x′ᵢ is YES, is x′ᵢ replaced randomly with another value. In Step 3, HM consideration, pitch adjustment or random selection is applied to each variable of the new harmony vector in turn.

In the case of combinational optimization problems with discrete representation or permutation encoding of solutions, the process of applying PAR parameter must be adapted accord to problem objective.

#### 4.1.4 Update Harmony Memory

If the new harmony vector x′=(x′₁,x′₂,…, x′ₙ ) is better than the worst harmony in the HM, judged in terms of the objective function value, the new harmony is included in the HM and the existing worst harmony is excluded from the HM.

#### 4.1.5 Check Stopping Criterion

If the stopping criterion (i.e. maximum number of improvisations) is satisfied, computation is terminated. Otherwise, Steps 3 and 4 are repeated.

### 4.2 Harmony Operators Accordance to Neural Oscillators Parameters

Neural oscillators parameters can be used directly with HS algorithm. Each solution corresponds to a TFS motion generator. Each solution in harmony memory, harmony vector, is encoded as an array of 10 elements, six elements for Matsouka oscillator variables and time periods and four other elemets for wights and bioses, where each element is a real variable.

#### 4.2.1 Initialization

For initialization, HM is filled with as many randomly generated solution vectors as the size of the HM (e.g. HMS). In the complete, each random generated vector is a valid neural oscillator parameter but for applying to robot, the probability that randomly generated neural oscillators are valid is low. So, alternatively, we develop an intelligence table for initialization (table 2). The lengths of intervals have been taken according to maximum and minimum degree of freedom of every joint of the robot. For example for knee joint angle, when τ₁ and τ₂ are between 0 and 25, the largeness of Matsuoka`s signal wavelength would be between 0 and 25, and by multiplying this signal by $w_{12}$, the whole search space would be covered, and by choosing $b_2$ between -5 and -125, this wavelength can generate signal by each virtual primal offset.

#### 4.2.2 Improvisation Step

New harmony vector, is generated based on process of part 3 in previous Section. Each element of harmony vector is selected from HM with probability HMCR and with probability (1 _ HMCR) is selected from set{1,2,…,10}.

In the pitch adjusting process, a value is set to a number in range from lower band to upper band with probability of PAR, or just stays in its original value with probability (1 _ PAR).

#### 4.2.3 Fitness function

TABLE 2
LOWER BAND AND UPPER BAND THAT USED FOR HARMONY MEMORY INITIALIZATION

| Symbol | Upper Bound | Lower Bound |
|---|---|---|
| $\tau_1$ | 25 | 0 |
| $\tau_2$ | 25 | 0 |
| α | 5 | -5 |
| β | 5 | -5 |
| γ | 5 | -5 |
| c | 2 | 4 |
| $w_{11}$ | 1 | -4 |
| $w_{12}$ | 0 | -5 |
| $b_1$ | 20 | -95 |
| $b_2$ | -5 | -125 |

Fitness function has a critical role in heuristic algorithms and is used to judge how well a solution represented by a vector is. To achieve more stable and faster walk, a fitness function based on robot's straight movement with having limited time for walking is assumed. The amount of deviation from straight walking is subtracted from the fitness as a punishment to force the robot to walk straight. The simulator runs for 15 seconds each time, during first 3 seconds robot is at lock phase. We assume when the robot is initialized each time, its X and Y values equal to zero and fitness function is calculated whenever robot falls or time duration for walking is over. Equation 9 shows the pseudo code of walking fitness function.

$$
\begin{aligned}
&\text{if } (CurrentTime \geq TimeDuration) \\
&\quad fitness = X - Y; \\
&\text{if } (RobotIsFallen) \\
&\quad fitness = \frac{X - Y}{TimeDuration - CurrentTime};
\end{aligned} \quad (9)
$$

According to equation 9 when robot falls down during walk the fitness is divided to remaining time of simulation. This punishment forces the robot to achieve a stable walk.

## 5 RESULTS

We ran the simulator on a Pentium IV 3 GHz Core 2.6 Duo machine with 4 GB of physical memory, with Linux SUSE 10.3 O.S. The time period for the simulation was 15 seconds. 2 hours after starting HS under the MATLAB, 750 trials were performed. Fig. 4 shows the average and best fitness values during these 750 trials.

Running the GA on the same system with 10 generations and 200 populations for each generation, after 7 hours results of figure 5 have been achieved.

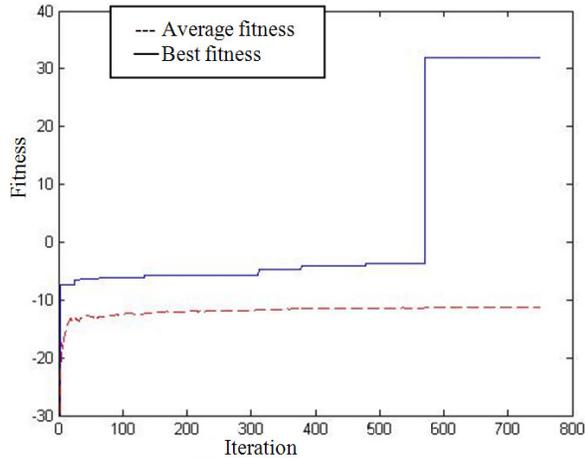

Fig. 4. HS Convergence. By HS as an optimizer the robot could walk 5.3 m in 12 s with average body speed 0.44.

Gait period at the best found fitness equals to 0.1 s. according to this consequence figures 6, 7 and 8, respectively show angular trajectories of hip, knee and ankle, generated by oscillators.

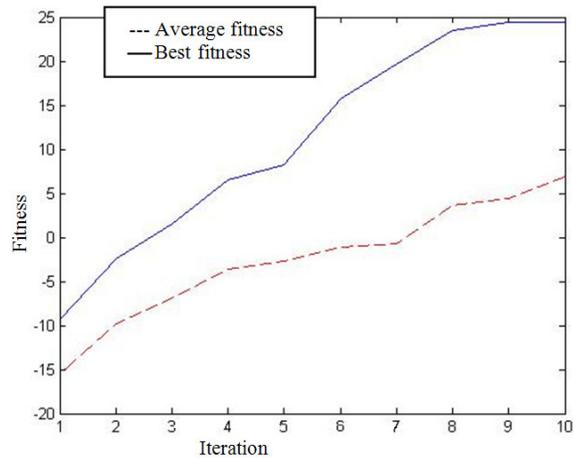

Fig. 5. GA Convergence. With respect to GA, By GA as an optimizer the robot only could walk 4.2 m in 12 s with average body speed 0.35.

Considerate towards the punishment set for changing direction of robot during movement, Fig. 9 shows the walking direction of the robot.

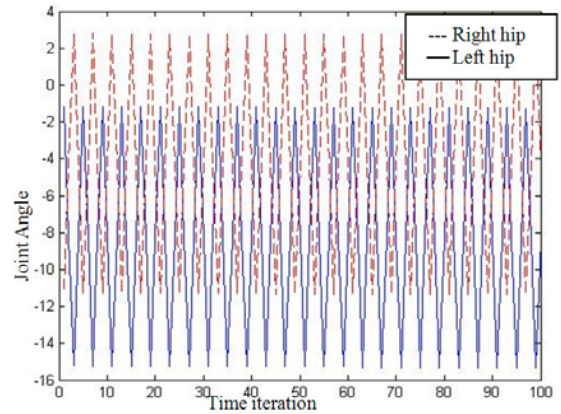

Fig. 6. left and right hip angle trajectories. Every 50 iterations equal to one second.

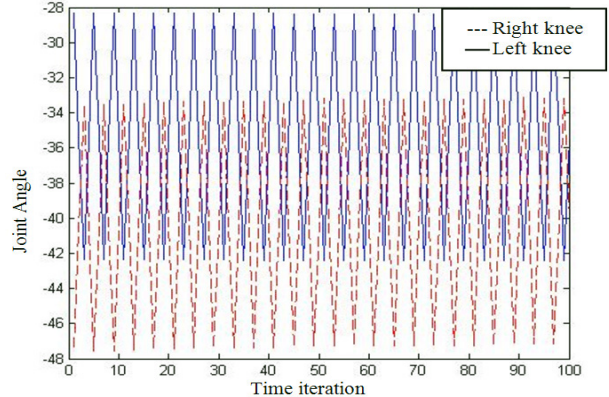

Fig. 7. left and right knee angle trajectories. Every 50 iterations equal to one second.

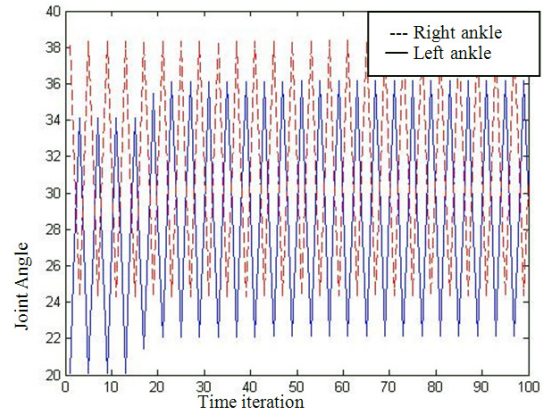

Fig. 8. left and right ankle angle trajectories. Every 50 iterations equal to one second.

According to the fact that during movement, pivot leg changes and this change causes changing human's hight, and we know that the ideal walking for a biped robot is a walking that is more similar to human walking. Thus, changes in height during walk are mandatory for a good movement. Figure 10 that illustrates the z-position of robot's torso during movement shows that our offering algorithm is much the same as human walking.

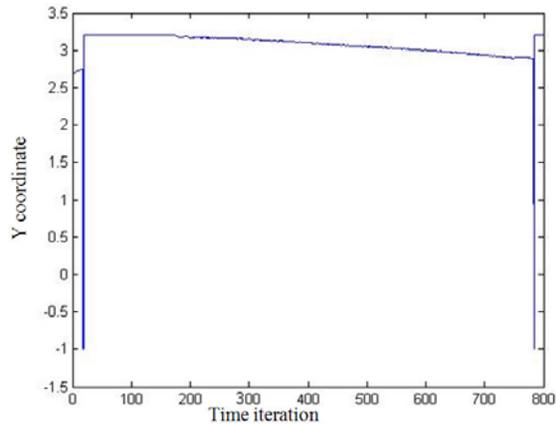

Fig. 9. this figure shows Y coordinate of robot during walk. Every 50 iterations equal to one second.

## 5 CONCLUSION

We have demonstrated the suitability of an evolutionary robotics approach to the problem of stable three-dimensional bipedal walking in simulation. The current implementation is capable of walking in a straight line on a planar surface with the use of proprioceptive input. When the neural network sends the signals to the robot, the input layer receives the feedback signals from hip of the robot on the other hand, in this study for the first time HS used for optimization of neural network weights. Using from HS as an optimizer shows that the harmony search algorithm is a computationally fast multi-objective optimizer tool for complex engineering multi-objective optimization problems. The main advantage of our walking controller model is having the least parameters compare with other gait generator models that it can helps the robot to find best solution sooner and sooner. It is also capable of being implemented on any kinds of humanoid robots without considering its physical model.

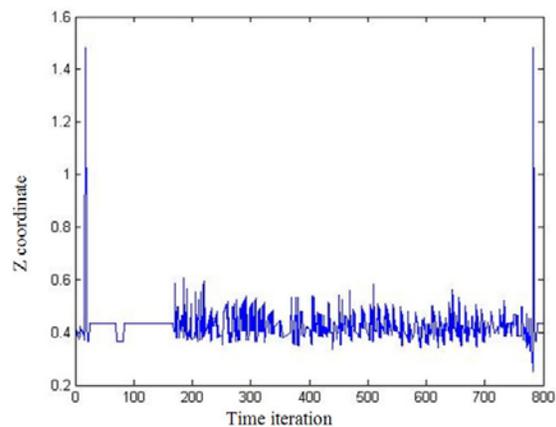

Fig. 10. Z-Position of robot's torso during movement. Every 50 iterations equal to one second.

**Ebrahim Yazdi** obtained his bachelor degree in Computer-software engineering from Qazvin Azad University, Qazvin, Iran. Currently, he is a master student in Department of computer-software engineering, Qazvin Azad University, Qazvin, Iran. He is a member of Mechateronic Research Lab. (MRL), Qazvin, Iran. He gots third place in Atlanta Robocup 2007 3DDevelopment league and gots first place in China Robocup 2008 3DDevelopment league.